\documentclass{llncs}
\usepackage{makeidx}  
\usepackage{subfig}
\usepackage{graphicx}
\usepackage{amssymb,amsmath,bm}
\usepackage{booktabs}
\usepackage[misc]{ifsym}
\usepackage{multirow}
\usepackage{bbding}
\usepackage{array}
\usepackage{float}
\usepackage{algorithm}
\usepackage{algorithmic}

\newcolumntype{C}[1]{>{\centering\let\newline\\\arraybackslash}m{#1}}
\begin{document}
	\frontmatter          
	%
	%
	\mainmatter              
	\title{Uncertainty-Guided Domain Alignment for Layer Segmentation in OCT Images}	
	\author{Jiexiang Wang\inst{1,2}, Cheng Bian\inst{2}, Meng Li\inst{3}, Xin Yang\inst{4}, Kai Ma\inst{2}, Wenao Ma\inst{1,2}, Jin Yuan\inst{3}, Xinghao Ding\inst{1},  Yefeng Zheng\inst{2}}	
	\institute{School of Information Science and Engineering, Xiamen University, China\and Tencent YouTu Lab \and  State Key Laboratory of Ophthalmology, Zhongshan Ophthalmic Center, Sun Yat-sen University, Guangzhou 510060, China \and Dept. of Computer Science and Engineering, The Chinese University of Hong Kong \\
		\email{wangjx@stu.xmu.edu.cn}, \email{tron1992@foxmail.com}, \email{limeng48@mail2.sysu.edu.cn}, \email{yangxinknow@gmail.com}, \email{kylekma@tencent.com}, \email{wenaoma@stu.xmu.edu.cn}, \email{yuanjincornea@126.com}, \email{dxh@xmu.edu.cn}, \email{yefengzheng@tencent.com}}
	\maketitle 
	\date{}
	\hyphenpenalty=5000
	\tolerance=1000
	
	\begin{abstract}
		Automatic and accurate segmentation for retinal and choroidal layers of Optical Coherence Tomography (OCT) is crucial for detection of various ocular diseases. However, because of the variations in different equipments, OCT data obtained from different manufacturers might encounter appearance discrepancy, which could lead to performance fluctuation to a deep neural network. In this paper, we propose an uncertainty-guided domain alignment method to aim at alleviating this problem to transfer discriminative knowledge across distinct domains. We disign a novel uncertainty-guided cross-entropy loss for boosting the performance over areas with high uncertainty. An uncertainty-guided curriculum transfer strategy is developed for the self-training (ST), which regards uncertainty as efficient and effective guidance to optimize the learning process in target domain. Adversarial learning with feature recalibration module (FRM) is applied to transfer informative knowledge from the domain feature spaces adaptively. The experiments on two OCT datasets show that the proposed methods can obtain significant segmentation improvements compared with the baseline models.
	\end{abstract}
	\section{Introduction}
\begin{figure}[t]
		\centering
		\setlength{\belowcaptionskip}{-0.3cm}
		\includegraphics[width=1.0\linewidth]{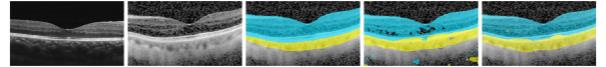}
		\caption{\small Peformance drop due to domain shift. First column: Original data from Optovue equipment. Second column: Original data from Heidelberg equipment. Third Column: The segmentation result of an Heidelberg OCT image using an established model trained on Heidelberg data. Fourth: The segmentation result of an Heidelberg OCT image using an established model trained on Optovue data. Fifth Column: The segmentation result of an Heidelberg OCT image using our model trained on Optovue data. The blue colors represents retinal layers, and the yellow colors represents choroidal layers.}
		\label{fig:performancedrop}
	\end{figure}

	Accurate segmentation for retinal and choroidal layers is fundamental to diagnose the progress of ocular diseases \cite{fujimoto2009optical}. For instance, glaucoma, high myopia, diabetic retinopathy, and even systemic diseases such as Alzheimer's disease, multiple sclerosis and obstructive sleep apnea-hypopnea syndrome, are related to retinal or choroidal lesions \cite{el2017correlation}. Furthermore, the changes of retinal thickness especially central retinal thickness reflect the response to therapy and have some relationship to vision prognosis in patients with macular edema. Recently, deep learning network has become a powerful tool for semantic segmentation on Optical Coherence Tomography (OCT) images \cite{sui2017choroid}. However, a well-trained segmentation model may underperform when being tested on data from different equipments or imaging protocols, which is caused by the domain shift (as shown in Fig.~\ref{fig:performancedrop}, performance drop mostly happened in boundary or junction regions, which were uncertain areas to segmentor). Many strategies like fine-tuning have been taken for the new domains in order to alleviate the performance drop. But it still needs massive data collection and enormous annotation workload which are impossible for many real-world medical scenarios. For this reason, constructing a general segmentation model suitable for various equipments is promising yet still challenging.
	Nowadays, Unsupervised Domain Adaptation (UDA) is an encouraging research direction that devotes to tackle the domain shift problem. Current research works such as CycleGAN \cite{zhu2017unpaired} and AdaptSegNet \cite{tsai2018learning} achieved significant improvement. In the medical imaging field, \cite{dong2018unsupervised} developed an UDA framework based on adversarial networks for lung segmentation on chest X-rays. \cite{ren2018adversarial} improved the UDA framework with Siamese architecture for Gleason grading of histopathology
	tissue. \cite{dou2018unsupervised} proposed a domain critic
	module (DCM) and a domain adaptation module (DAM) for the unsupervised cross-modality adaptation problem. These methods were based on the adversarial learning requiring empirical feature selection. Different from the above methods, \cite{zhang2017curriculum} presented a new approach by introducing curriculum-style learning on UDA, which was proved to be effective. \cite{zou2018unsupervised} employed the class-balanced self-training (CBST) framework to tackle the UDA problem by training the model with an easy-to-hard selection strategy in the target domain. However, curriculum-style methods might encounter the confirmation bias \cite{tarvainen2017mean} due to their definition on the "easy" task with high probability distribution. To measure the model confidence, Bayesian deep learning reveal that model uncertainty is indispensable. \cite{kendall2015bayesian} showed that there is a strong inverse correlation between class accuracy and model uncertainty. Although there have been many works adopting deep learning to estimate the uncertainty \cite{kendall2015bayesian}\cite{sedai2018joint}\cite{nair2018exploring}, they have not utilized it in the training stage.  
	
	In this work, we propose an uncertainty-guided domain alignment method for the UDA task, which help the established model segment the retinal and choroidal layers accurately in the target domain. Our contribution can be summarized into four folds: first, inspired by \cite{kohl2018probabilistic}, we use a simply end-to-end network to estimate the uncertainty, which makes it possible for a joint uncertainty training. Second, an uncertainty-guided cross-entropy loss is designed to enhance the transfer ability of our model over areas with high uncertainty. Third, we develop an uncertainty-guided curriculum transfer strategy for self-training (ST) to progressively enhance the performance in the target domain with the enlarged data having low uncertainty. Lastly, we apply adversarial learning with a feature recalibration module (FRM) inspired by \cite{roy2018concurrent} to intergrate multi-level features without manual selection. The proposed method and its variants are extensively evaluated on two datasets, one of which includes 623-patient OCT images obtained from Heidelberg devices, and the other one contains 537-patient OCT images obtained from Optovue devices.
	\vspace{-0.5cm}
	\begin{figure}[H]
		\centering
		\subfloat{\includegraphics[width=1.0\textwidth]{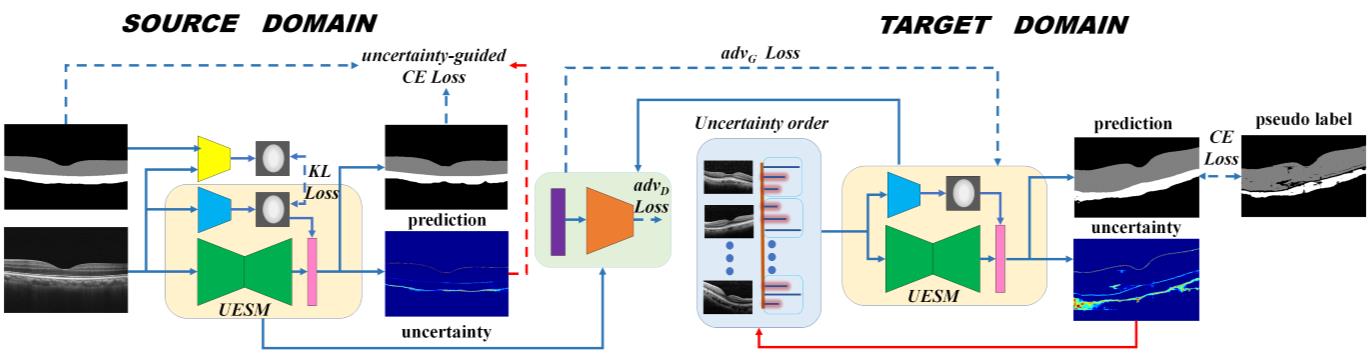}}
		\caption{\small The workflow of the proposed method: Green blocks: PSPNet. Pink blocks: prediction network (Fig. 3(c)). Blue blocks: prior network. Yellow blocks: posterior network. Purple block: FRM. Orange block: discriminator.} 

	\end{figure}
	\vspace{-1.0cm}
	\section{Method}
	Fig. 2 overviews our segmentation method for retinal and choroidal layers in OCT images. We use ResNet18 with PSP module (PSPNet) (Fig. 3(a)) as our backbone architecture \cite{zhao2017pyramid}. To estimate the uncertainty based on the variance of multiple outputs, a prior network $M_{prior}$ (Fig. 3(b)) and a posterior network $M_{post}$ (Fig. 3(b)) are adopted to model the complex distributions. Moreover, the feature recalibration module (FRM) (Fig. 3(d)) is applied to transfer multi-level feature information.  

	\subsection{Segmentation and Uncertainty Estimation}
	The core concept of the proposed method is to train a deep learning network capable of simultaneously estimating the segmentation result for an input image and giving an uncertainty measurement, which provides additional information besides the deterministic outcomes. Different from the existing methods using dropout over spatial features to estimate the uncertainty \cite{kendall2015bayesian}\cite{sedai2018joint}\cite{nair2018exploring}, we adopt an efficient solution from \cite{kohl2018probabilistic} that combines a segmentation network with a conditional variational autoencoder (CVAE) for the uncertainty estimation, which is shown in Fig. 2. We refer this architecture as uncertainty estimation and segmentation module (UESM) for convenience reasons.\par
	The uncertainty measurement is based on the variance of the network’s output. To generate diverse predictions from a given input image, UESM utilize a pre-trained prior network ($M_{prior}$) to process the input image $N$ times and draws segmentation samples from a low-dimentional latent space that covers all presented segmentation variants. The sampling process from the latent space is mathematically formulated as:
	
	\begin{small}
		\begin{equation}
		z_{i} \sim D_{prior}(\cdot|X) = \mathcal N(\mu_{prior}(X;\Theta_{prior}),diag(\sigma_{prior}(X;\Theta_{prior})))
		\end{equation}
	\end{small}where $X$ is OCT images; $D_{prior}(\cdot|X)$ is the prior distribution modelled as a multivariate Gaussian distribution; $\mu_{prior}$ and $\sigma_{prior}$ are the mean and variance of $D_{prior}(\cdot|X)$. To train such prior network that represents the informatic latent space with the prior distribution, we use a posterior net that learns from the given input images and corresponding ground truth to simulate a posterior distribution. The Kullback-Leibler divergence $D_{KL}(D_{post}(z|Y_{s},X_{s})||D_{prior}(z|X_s))$ is applied to penalize differences between the two distributions in the source domain, where $X_{s}$ are source OCT images; $Y_{s}$ are the corresponding source OCT annotations; $D_{post}$ is the posterior distribution modelled as a multivariate Gaussian distribution. Additionally, the prior net offers feature maps that are concatenated with the segmentation network’s output to generate the final segmentation mask, so the cross-entropy loss is also used to penalize differences between the segmentation output and groundtruth. More training details of the UESM can be found in \cite{kohl2018probabilistic}.\par
	Assume that we are given an OCT image $x$, the sampling process repeats $N$ times to obtain $N$ different features from the prior distribution, which are concatenated with the extracted features from the segmentation network to form a series of Monte Carlo samples ${\hat{Y}_{1}, ..., \hat{Y}_{N}}$, and the uncertainty is estimated by calculating the variance of the final segmentation outputs \cite{nair2018exploring}. Finally, we obtain the mean prediction $Y(x)$ and mean uncertainty $U(x)$ by
	\begin{small}
		\begin{equation}
		Y(x)=\frac{1}{N}\sum_{n=1}^{N}{\hat{Y}_{n}}
		\end{equation}
		\vspace{-10pt}
		\begin{equation}
		U(x)=\frac{1}{C}\sum_{c=1}^{C}var(\hat{Y}_{1}, ..., \hat{Y}_{N}).
		\end{equation}
	\end{small}
	\subsection{Uncertainty-guided cross-entropy loss for UDA}
	Since the labels for source domain are available, we train the network with a modified cross-entropy loss. Different from the vanilla weighted cross-entropy loss, which handles unbalanced data with different weights, we design an uncertainty-guided cross-entropy loss for boosting the performance over areas with high uncertainty in the source domain. Synergistic training between our designed loss function and adversarial learning can also enhance the transfer ability of our model over areas of high uncertainty in target domain. The designed loss function is formulated as
	\begin{small}
		\begin{equation}
		\mathcal L^{s}(X_{s},Y_{s};\Theta_{g})=-\mathbb{E}_{x_{s}\sim S}(\sum_{i=1}^{N_{s}}\sum_{c=1}^{C}y_{s,i,c}\log G(x_{s,i};\Theta_{g})) \odot (\mathbf{1}+Normalize(U_{x_{s}})).
		\end{equation}
	\end{small}For each source OCT image $x_{s}$, there is one corresponding annotation $y_{s}$; $G$ is UESM; $\Theta_{g}$ is the parameters of $G$; Symbol $\odot$ represents the element-wise multiplication. $U_{x_{s}}$ represents the uncertainty map of $G$. $\mathbf{1}$ represents an all one metrix as the size of $U_{x_{s}}$. Since the absolute uncertainty values $U_{x_{s}}$ can be small, we normalize the value to [0, 0.1].

    \subsection{Uncertainty-guided self-training for UDA}
    We develop an uncertainty-guided curriculum learning for self-training to make the model better adapted to the target domain. Different from other methods that distinguish task difficulties with a softmax probability, we develop a novel easy-to-hard learning strategy based on the uncertainty. In each training epoch, we maintain a sorted list based on the uncertainty values $U$ and the corresponding pseudo label list $Y$. Data close to the head of the list is the one with higher certainty, while the latter ones are the hard cases. In the curriculum learning for self-training process, we train our model with the enlarged training-set with vanilla cross-entropy loss $\mathcal L^{t}(X_{t},\hat{Y}_{t};\Theta_{g})$ and update the network using an easy-to-hard learning strategy.\par
 	\vspace{-0.5cm}   
	\begin{figure}[H]
		\centering
		\includegraphics[width=1.0\textwidth]{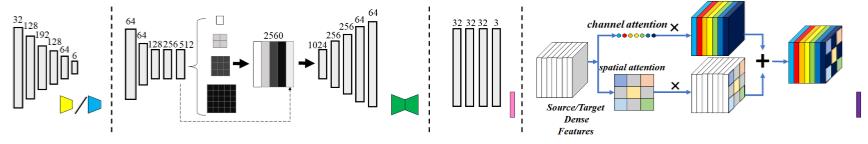}
		\vspace{-5pt}
		\caption{\small Architecture of the sub-networks in our framework: (a) The PSPNet; (b) The prior/posterior network; (c) The prediction network; (d) The feature recalibration module (FEM).}
		\label{fig:FRM}
	\end{figure}
	\vspace{-1.0cm}
	\subsection{Adversarial Learning with FRM}
	In the target domain, due to the lack of annotations, we leverage the adversarial learning to train the UESM by minimizing the discrepancy in the feature space between the source and target domains.  Similarly, we use the PatchGAN as our discriminator \cite{zhu2017unpaired}. Different from \cite{dong2018unsupervised}\cite{ren2018adversarial}\cite{dou2018unsupervised} only using high-level features for UDA, our adopted FRM can leverage the full feature spectrum and automatically select prominent features. We use the pre-trained UESM with source domain data to extract the features of target domain data. In the UESM, each feature scale generates one output feature map in the same dimension via convolution and upsampling operations. The feature maps are further processed by the channel-wise attention and spatial-wise attention modules to highlight the prominent features and suppress the irrelevant ones. The combined feature maps are then fed to the discriminator network for the adversarial learning, whose loss can be defined as
	\begin{small}
		\begin{equation}
		\mathcal L_{adv_{D}}=-\mathbb{E}_{x_{s}\sim S}\log D(R(G(x_{s};\Theta_{g});\Theta_{r});\Theta_{d})-\mathbb{E}_{x_{t}\sim T}(1-\log D(R(G(x_{t};\Theta_{g});\Theta_{r});\Theta_{d}))
		\end{equation}
		
		\begin{equation}
		\mathcal L_{adv_{G}}=-\mathbb{E}_{x_{t}\sim T}\log D(R(G(x_{t};\Theta_{g});\Theta_{r});\Theta_{d})
		\end{equation}
	\end{small}where $x_{s}$ is source OCT data; $R$ is the FRM; $\Theta_{r}$ is the parameters of $R$; $D$ is the discriminator; $\Theta_{d}$ is the parameters of $D$.\par
	Combined with the aforementioned loss, the full objective function  
	\begin{small}
		\begin{equation}
		\mathcal L_{FULL}=\lambda_{s}\mathcal L^{s}+\lambda_{t}\mathcal L^{t}+\lambda_{D}\mathcal L_{adv_{D}}+\lambda_{G}\mathcal L_{adv_{g}}+\lambda_{KL}D_{KL}.
		\end{equation}
	\end{small}
	where $\lambda_{s}$ is the weight for $\mathcal L^{s}$; $\lambda_{t}$ is the weight for $\mathcal L^{t}$; $\lambda_{D}$ is the weight for $\mathcal L_{adv_{D}}$; $\lambda_{G}$ is the weight for $\mathcal L_{adv_{g}}$; $\lambda_{KL}$ is the weight for $D_{KL}$.

	\vspace{-0.2cm}
	\section{Experiment}
	\paragraph{\textbf{Dataset.}}
	The validation of the proposed method is performed in a local clinical dataset covering 1160 patients. In our dataset, the 537 source domain images are obtained from AngioVue OCT system (Optovue, Inc., Freemont, CA) while the 623 target domain images are gathered from Heidelberg Spectralis OCT instrument(Heidelberg Engineering, Inc., Heidelberg, Germany). All slices provided were grayscale with the size of 630$\times$496 pixels. Experienced experts annotate the retinal layers and choroidal layers manually and elaborately as ground truth. To better demonstrate the effectiveness and intuitive of our approach, both the source and target images without performing any image pre-processing operation. 
	
	\paragraph{\textbf{Implementation Details.}}
	In our experiments, we initialize the uncertainty estimation and segmentation module parameters with source domain datasets using an Adam optimizer (batch size=1, initial learning rate is 0.001 and total iteration=50000). During the domain adaptation, we update the weights of all sub-networks using smaller initial learning rate 0.0001 with 36000 iterations. We set $N$ to 4 to balance between training time and uncertainty quality. $\lambda_{s}$ is set to 1.0, $\lambda_{t}$ is set to 0.1, $\lambda_{R}$ is set to 0.003, $\lambda_{D}$ is set to 1.0. We implementate our whole network with PyTorch, using a standard PC with an NVIDIA Tesla P40. 
	\begin{figure}[h]
		\centering
		\subfloat[ ]{\includegraphics[width=0.19\textwidth]{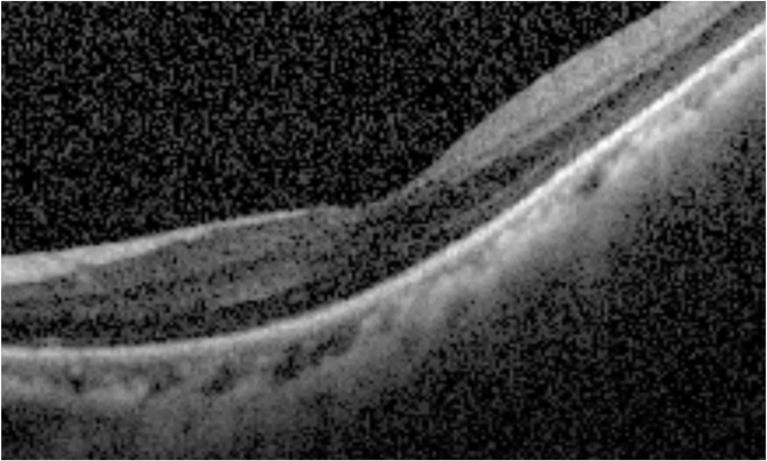}}	
		\subfloat[ ]{\includegraphics[width=0.19\textwidth]{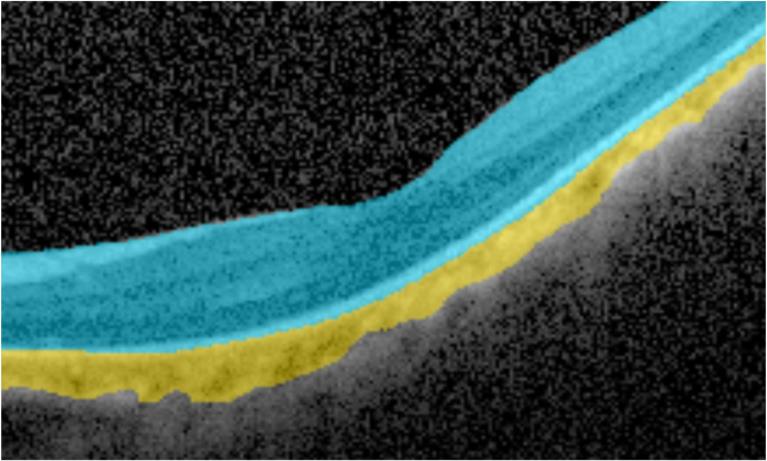}}
		\subfloat[ ]{\includegraphics[width=0.19\textwidth]{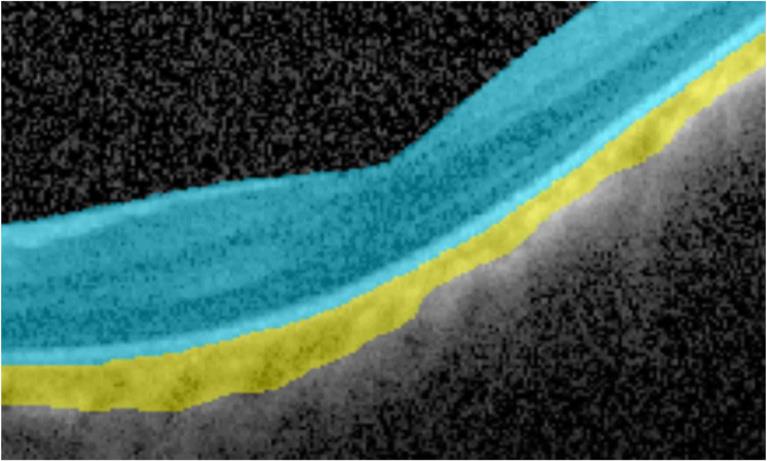}}
		\subfloat[ ]{\includegraphics[width=0.19\textwidth]{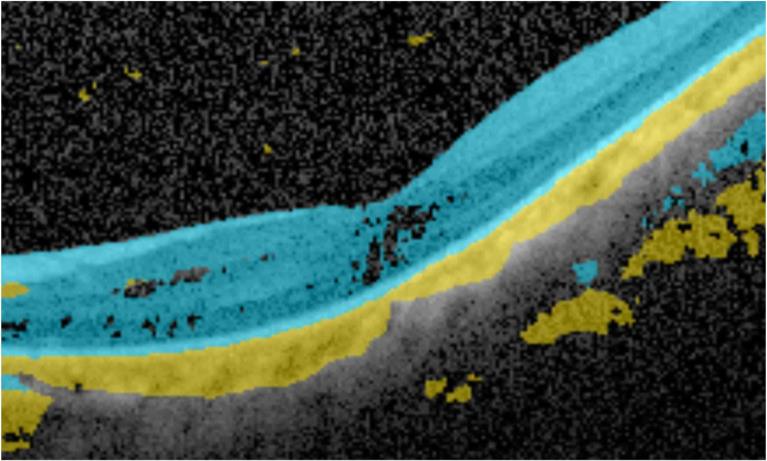}}
		\subfloat[ ]{\includegraphics[width=0.19\textwidth]{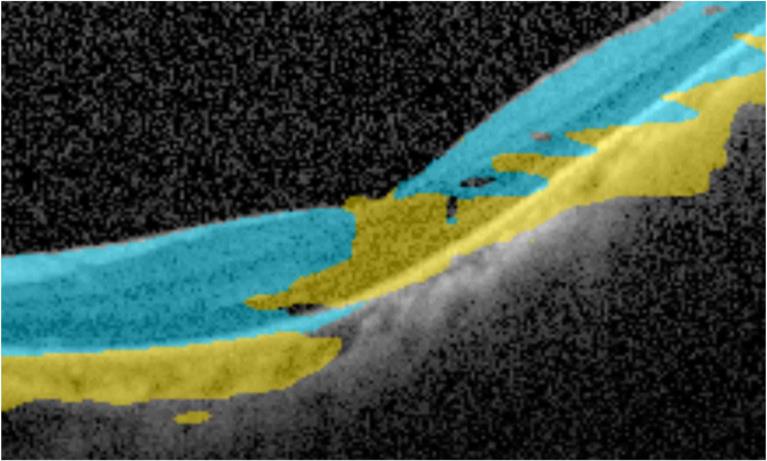}}\\
		\vspace{-10pt}
		\subfloat[ ]{\includegraphics[width=0.19\textwidth]{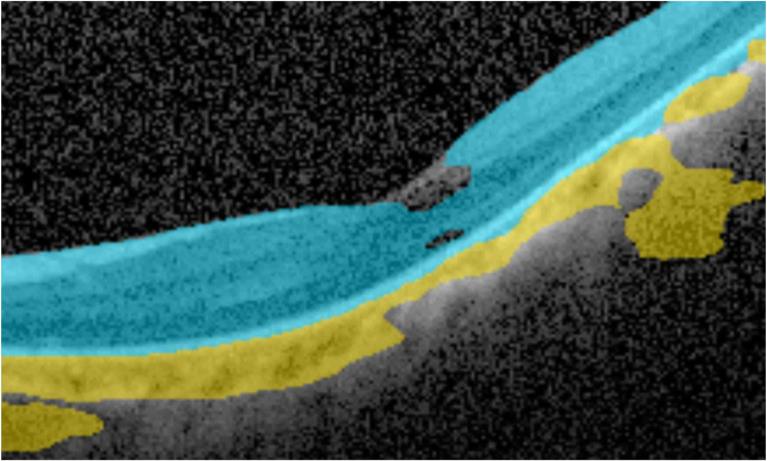}}
		\subfloat[ ]{\includegraphics[width=0.19\textwidth]{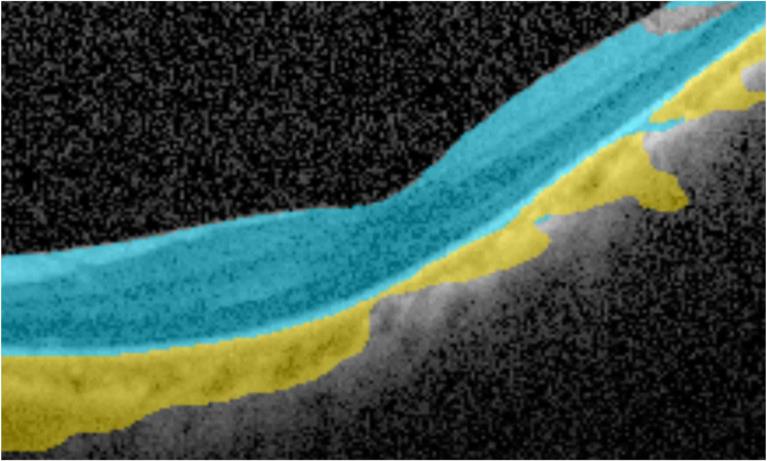}}
		\subfloat[ ]{\includegraphics[width=0.19\textwidth]{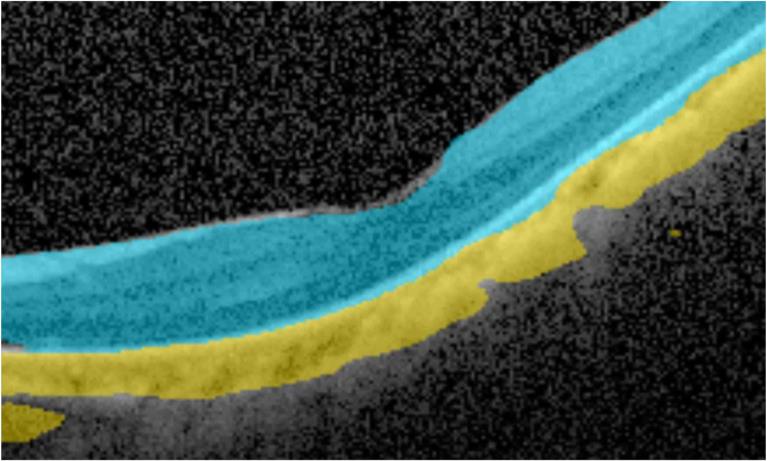}}	
		\subfloat[ ]{\includegraphics[width=0.19\textwidth]{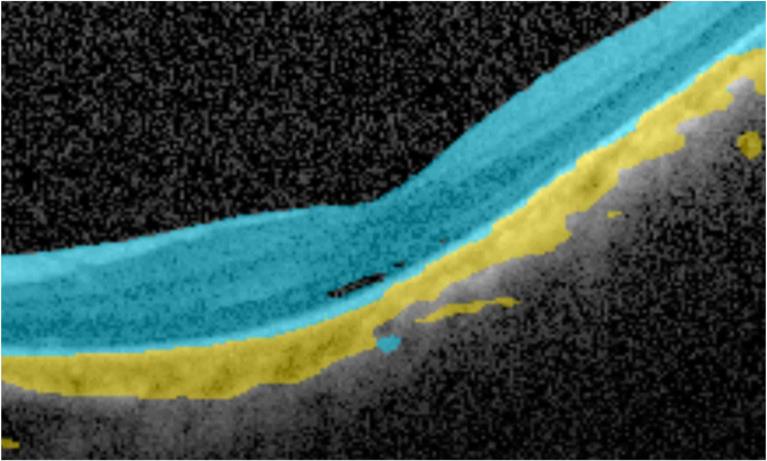}}	
		\subfloat[ ]{\includegraphics[width=0.19\textwidth]{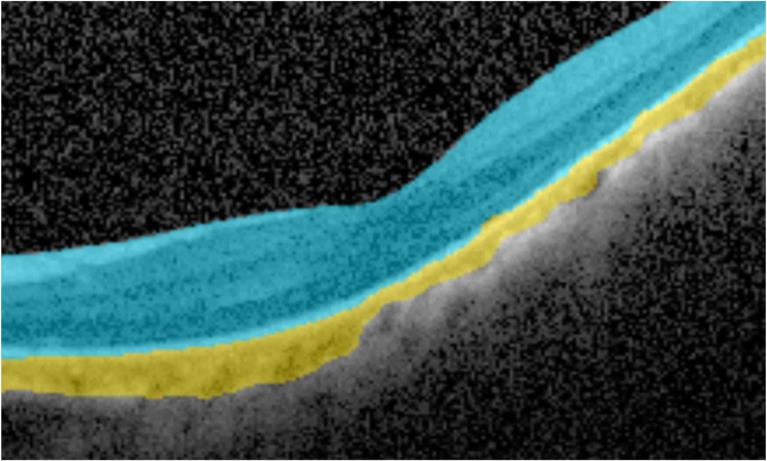}}\\
		\vspace{-5pt}
		\caption{\small Visual comparison for the retinal and choroidal segmentation results from different state-of-art algorithms and ablation setting. (a) Original image from  $T$. (b) annotation. (c) Orig T2T. (d) Orig S2T. (e) CycleGAN. (f) AdaptSegNet. (g) Orig S2T+ADV. (h) Orig S2T+ADV+FRM. (i) Orig S2T+ADV+FRM+UCE. (j) Orig S2T+ADV+FRM+UCE+UST. }
		\label{fig:VisualComparison}
	\end{figure}	

	\paragraph{\textbf{Quantitative and Qualitative Analysis.}}
	In order to clarify the effectiveness of the proposed framework, we adopt Dice cofficient (DSC), Conformity (conf) for further evaluation. We firstly trained two PSPNet on the source OCT data (Orig S2T) and the target OCT data (Orig T2T) respectively with the same settings, and then test them on the target OCT data. The results in Table 1 shows that the mean Dice in Orig S2T drop about $10.043\%$ than Orig T2T. As we can see, our method can promote about $7.333\%$ in Dice than Original S2T. From another aspect, our framework performance extremely reaches to the Orig T2T which proves the efficacy of the proposed method in UDA problem.\par
	\begin{table}[H] \caption {Quantitative evaluation of our proposed methods} \label{table:quanti_metric}
		\scriptsize
		\centering
		\begin{tabular}{l|C{1cm}|C{1.3cm}|C{1cm}|C{1.3cm}|C{1cm}|C{1.3cm}}			
			\toprule[2pt]
			\multirow{2}{*}{\bf{Method}} & \multicolumn{2}{C{2cm}|}{\bf{Metrics on Retinal}} & \multicolumn{2}{C{2cm}|}{\bf{Metrics on Choroidal}} & \multicolumn{2}{C{2cm}}{\bf{mean}}\\
			\cline{2-7}		
			&Dice[\%]$\uparrow$ 	&Conf[\%]$\uparrow$  	 	&Dice[\%]$\uparrow$ 	&Conf[\%]$\uparrow$   	&Dice[\%]$\uparrow$ 	&Conf[\%]$\uparrow$\\
			\cline{0-1}
			\cline{2-7}
			Orig T2T &97.954 &95.822  &91.829 &82.203 &94.892 &89.013\\
			\cline{0-1}
			\cline{2-7}
			Orig S2T &90.149 &78.147  &79.548 &48.580 &84.849 &63.364\\
			CycleGAN~\cite{zhu2017unpaired} &88.479 &73.958  &77.053 &40.438 &82.766 &57.198\\
			AdaptSegNet~\cite{tsai2018learning} &94.389 &88.110 &84.549 &63.452 &89.469 &75.781\\
			\cline{0-1}
			\cline{2-7}
			
			Orig S2T+ADV &95.193 &89.899  &85.733 & 66.718 &90.463 &78.309\\
			Orig S2T+ADV+FRM &95.349 &90.244  &86.427 &68.594 &90.888 &79.419\\
			Orig S2T+ADV+FRM+UCE &95.641 &90.886  &87.039 &70.219 &91.340 &80.553\\
			Orig S2T+ADV+FRM+UCE+UST &\textbf{96.065} &\textbf{91.808} & \textbf{88.298} &\textbf{73.494} &\textbf{92.182} &\textbf{82.651}\\
			\hline
			\toprule[2pt]
		\end{tabular}
	\end{table}
	\vspace{-0.5cm}
	In further experiment, we compare our approach with state-of-the-art domain adaptation approaches including CycleGAN \cite{zhu2017unpaired} and AdaptSegNet \cite{tsai2018learning}. Compared to CycleGAN and AdaptSegNet, experiments show our method outperforms the state-of-art algorithms by $9.416\%$ and $2.713\%$ in UDA problem.\par
			
	\vspace{-0.5cm}
	\begin{figure}[h]
		\centering
		\hspace{1pt}
		\subfloat[]{\includegraphics[width=0.185\textwidth]{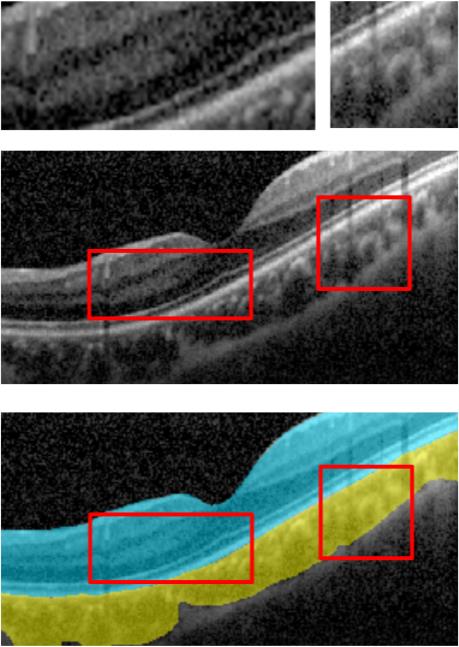}}
		\hspace{1pt}
		\subfloat[]{\includegraphics[width=0.185\textwidth]{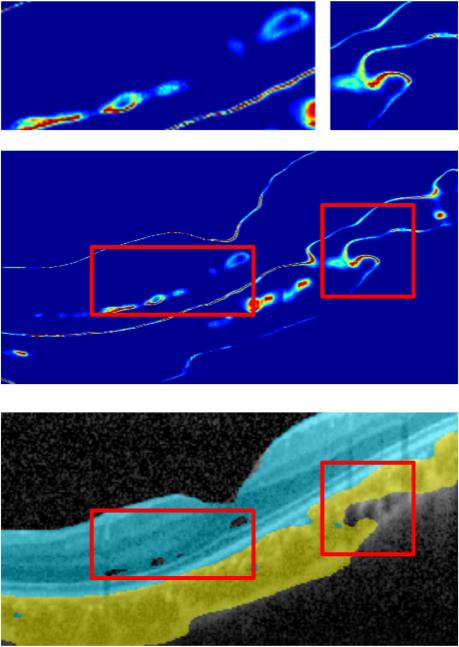}}		
		\hspace{1pt}
		\subfloat[]{\includegraphics[width=0.185\textwidth]{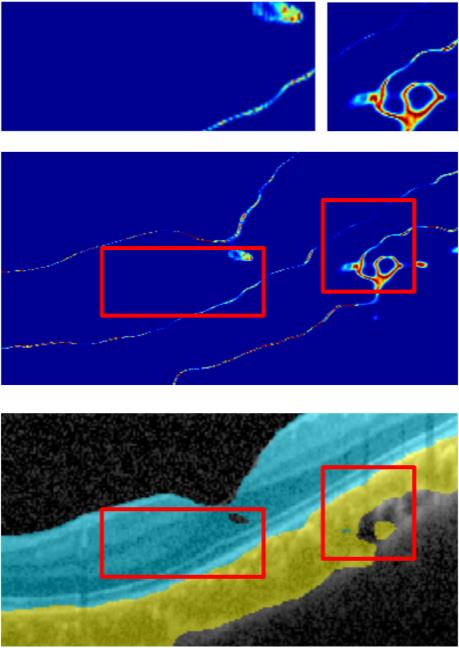}}	
		\hspace{1pt}
		\subfloat[]{\includegraphics[width=0.185\textwidth]{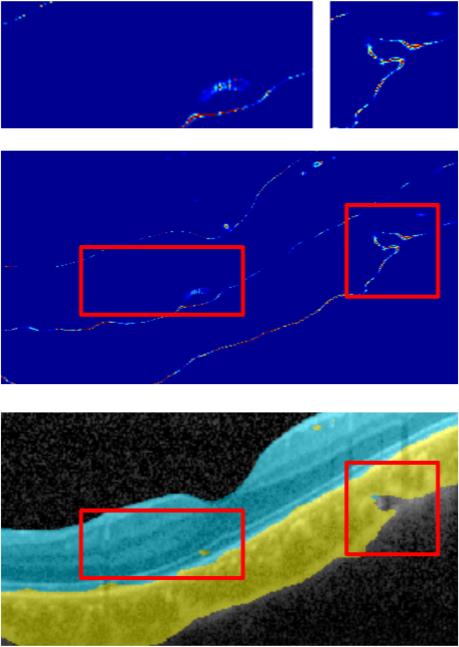}}
		\hspace{1pt}
		\subfloat[]{\includegraphics[width=0.185\textwidth]{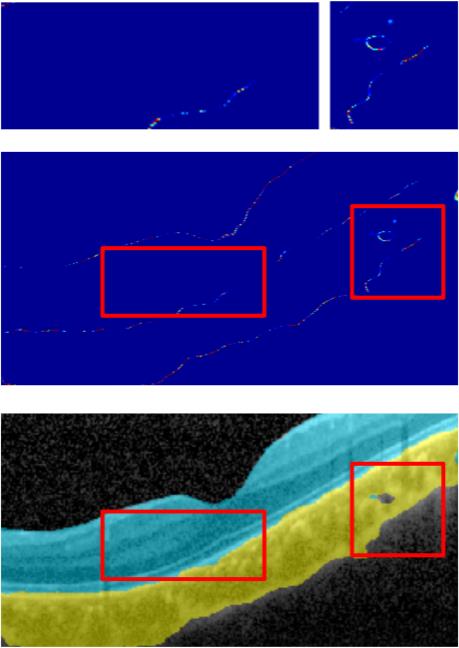}}
		\vspace{-5pt}
		\caption{\small The model uncertainty decreased during training stage: (a) The reference image. (b) 0 iterations. (c) 10000 iterations. (d)20000 iterations. (e) 30000 iterations.}
		\label{fig:Uncertainty}
	\end{figure}
	\vspace{-0.5cm}
	In addition, we examine the effect of the feature recalibration module (FRM), uncertainty-guided cross-entropy loss (UCE) and uncertainty-guided curriculum learning for self-training (UST) on the performance in the target domain. Orig S2T+ADV means our method without FRM, UCE and UST. The result of the ablation study in Table \ref{table:quanti_metric} shows that our proposed module can achieve a better Dice (up to $1.719\%$) than Orig S2T+ADV. On the other hand, Fig.~\ref{fig:VisualComparison} demonstrate that every proposed modules can contribute to alleviate the domain misalignment and outperforms the state-of-the-art algorithms. Moreover, Fig.~\ref{fig:Uncertainty} shows our method can reduce the uncertainty in the training stage progressively, which also indicates the rising performance in the target domain gradually.

	\section{Conclusion}
	In this paper, we proposed an uncertainty-guided domain alignment method for retinal and choroidal layers segmentation in OCT images. Our method innovatively integrates uncertainty into the training stage in UDA problem, which not only enhances the model transfer ability over areas of uncertainty, but also promotes the target domain performance progressively. Also, the applied adversarial learning with feature recalibration module (FRM) can enhance the target domain performance sigficantly.

	%
	\bibliography{uncertainty}
\end{document}